# Bayesian Inference in Monte-Carlo Tree Search

**Gerald Tesauro, VT Rajan, and Richard Segal**
IBM TJ Watson Research Center
Yorktown Heights, NY 10598

## Abstract

Monte-Carlo Tree Search (MCTS) methods are drawing great interest after yielding breakthrough results in computer Go. This paper proposes a Bayesian approach to MCTS that is inspired by distribution-free approaches such as UCT [13], yet significantly differs in important respects. The Bayesian framework allows potentially much more accurate (Bayes-optimal) estimation of node values and node uncertainties from a limited number of simulation trials. We further propose propagating inference in the tree via fast analytic Gaussian approximation methods: this can make the overhead of Bayesian inference manageable in domains such as Go, while preserving high accuracy of expected-value estimates. We find substantial empirical outperformance of UCT in an idealized bandit-tree test environment, where we can obtain valuable insights by comparing with known ground truth. Additionally we rigorously prove on-policy and off-policy convergence of the proposed methods.

## 1  INTRODUCTION

*Monte-Carlo Tree Search* (MCTS) methods provide a means of on-the-fly planning in complex sequential decision problems. MCTS couples traditional tree search techniques with node evaluations that are based on stochastic simulations, i.e., "playouts." The use of MCTS has led to spectacular results in the domain of computer Go: within a few years, the best programs have gone from weak intermediate play to grandmaster level in $9 \times 9$ Go, and expert level in $19 \times 19$ Go. MCTS also soundly defeated traditional alpha-beta search programming in the recent 2008 General Game-Playing competition held at AAAI [10]. As a result, there is now intense interest in further development of MCTS methods, and using them in numerous other planning and gaming applications [6, 14, 16].

MCTS algorithms were first developed specifically for computer Go [4, 9]. More recently, a number of general and theoretically principled algorithms have been developed, including UCT [13], BAST [8] and HOO [5]. These algorithms build upon seminal work on the UCB1 bandit sampling algorithm [2], and focus on providing strong worst-case distribution-free convergence guarantees and bounds on the cumulative regret. They also have the further merit of very low computational complexity. The great results obtained using UCT in the French program MoGo [11] and several other Go programs are probably most reponsible for the current excitement about MCTS methods.

In contrast to the above-mentioned distribution-free algorithms, we propose in this paper a fundamentally different approach based on Bayesian inference. The basic premise of our work is that, in practical applications of MCTS to MDPs or games, developers will usually know detailed characteristics of the reward distributions in the domain. In fact, we note that current applications of MCTS envision trials that are performed exclusively in simulation; an accurate reward model is thus a prerequisite to performing faithful simulated MCTS trials. Algorithms making use of such reward models could potentially outperform distribution-free methods that assume no knowledge of the reward distributions.

In our proposed Bayesian MCTS approach, stochastic trial results at leaf nodes are combined with prior reward information to yield posterior distributions; these then propagate upward according to the appropriate inference model to determine interior node distributions. In MDPs a parent's distribution is obtained by applying a distributional MAX operator to its child node distributions (as we detail in section 3), while in two-player zero-sum games, parent nodes use MAX or MIN operators according to odd/even node depths. If the leaf node priors and inference models are correct, this methodology can enable Bayes-optimal estimation of interior node values. As we show in section 5, such estimates can be much more accurate for a limited number of trials than the analogous UCT estimates, which are based on a simple average-value backup. As noted in [9], average-value estimates are "very inefficient," and often significantly underestimate the true expected-max value of parent nodes. This problem becomes especially pronounced in trees with non-uniform tree width, and we find that Bayesian MCTS can very substantially outperform UCT (by more than an order of magnitude) in this case.

We do recognize that Bayesian inference entails greater computational cost than existing MCTS methods, and therefore does not automatically provide a net win in terms of quality of solution for a given amount of simulation CPU time. However, as we discuss in section 4.1, recent advances in fast analytic Gaussian approximation of MIN/MAX distributions reduce Bayesian MCTS compute times to within an order of magnitude of UCT. This suggests that Bayesian MCTS can be effective in applications where the cost of executing the simulation steps greatly exceeds the time needed to compute the MCTS sampling decisions. There are many examples of such compute-intensive simulation domains, e.g., in routing and scheduling in large-scale networks, drug design, and computational physics. The game of Go also has this property – MCTS trials in $19 \times 19$ Go typically descend the tree via bandit sampling for $\sim O(10)$ steps to reach a leaf node, which is then evaluated by executing a few hundred steps of a stochastic "playout policy" until a terminal state is reached. In addition, we expect that further significant speedups can be obtained by developing incremental versions of these approximation methods; this is a topic of our ongoing and future work.

A final potential advantage of Bayesian MCTS is that it can allow more robust convergence under a wide range of sampling policies. The convergence of UCT relies on focusing the vast majority of trials on the optimal path. By contrast, it is intuitively clear that Bayesian MCTS will converge to correct minimax values even with uniform random sampling. In section 3.1 we provide sketches of rigorous proofs of both on-policy and off-policy convergence of our Bayesian version of UCT. We expect that off-policy convergence will enable more robust convergence in massively parallel implementations of MCTS where it would be difficult to achieve UCB1's focusing effect.

While probabilistic inference has been previously studied in game-tree contexts [15, 3], ours is the first study to treat uncertainties arising from stochastic sampling. Moreover, despite the title of [3], we believe our work is the first truly "Bayesian" approach to minimax trees in the strict sense of combining prior distributions with sample evidence to compute posterior distributions.

## 2   UCB/UCT BACKGROUND

Algorithm UCB1 [2] elegantly solves the exploration-exploitation dilemma in the multi-armed bandit problem. It maintains counts $\{n_i\}$ of the number of trials of each arm, along with the average rewards $\{\bar{r}_i\}$ obtained in those trials, where rewards are scaled to lie in $[0, 1]$. It then computes at each time step an upper confidence bound $B_i$ for each arm $i$ according to:

$$B_i = \bar{r}_i + \sqrt{2 \ln N / n_i} \qquad (1)$$

where $N = \sum n_i$ is the total trials of all arms. The arm with maximum bound $B_i$ is then selected for the next trial. (Any nodes with $n_i = 0$ are considered to have an infinite bound and thus would be preferred to nodes that have already been tried at least once.) UCB1 is proven to converge to selecting the optimal arm with probability 1 as $N \to \infty$, while continuing to sample inferior arms at a logarithmic rate. Also, its cumulative regret achieves optimal scaling $\sim \ln N$.

UCT [13] employs UCB1 in the context of multi-stage decision problems, where a generative model is available to simulate the problem. The algorithm is intended to apply to a variety of problems, including games with strict tree structures, as well as more general MDPs allowing trajectories containing loops. Consequently, [13] does not give a precise specification regarding state space representation, or how nodes are added to the tree/state space over time. The only characteristics that are precisely specified are: (1) a number of simulated trials are launched from a common start state/node, and average reward statistics are maintained for all states/nodes in the representation that are encountered during the trials; (2) the decision at every step is made by maximizing the UCB1 bound formula of equation 1.

A common implementation of UCT in two-player games such as Go makes use of the UCB1 policy plus an additional "playout policy." Trials begin at the root node, and descend through the tree according to UCB1, until a node is encountered where there are some child moves not yet represented in the tree nodes. At that point, a new a leaf node is created, and play continues using the playout policy until a terminal game state is reached. The trial result is added to the average reward statistics for every tree node in the path of the trial, and the next trial starts again at the root node. Due to the selective effect of UCB1, trials gradually focus on the minimax optimal line of play, with the result that the average reward values eventually converge to the correct game-theoretic values.

We note that the exploration term in UCB1 is analogous to algorithms such as Interval Estimation [12] which are explicitly based on estimating the uncertainty in the value of a given state. By that analogy, the $1/\sqrt{n_i}$ factor in UCB1 exploration may be viewed as a crude (under)estimate of the uncertainty in a UCT node value. A much more informed uncertainty estimate could be obtained based on the number of child nodes, and their associated values and uncertainties.

## 3 BAYESIAN MCTS FORMALISM

We now present our Bayesian reformulation of MCTS, comprising mechanisms for computing leaf node posterior distributions, propagating inference up the tree to compute interior node distributions, and then computing distribution-based upper confidence estimates as a basis for choosing where to sample next. Our formal description here is limited to the case of uncorrelated payoffs at leaf nodes, although the Gaussian approximation framework of section 4.1 does permit usage of general correlation matrices $\rho_{ij}$ between pairs of sibling node distributions. We further assume a static set of leaf nodes where prior/posterior calculations are performed. We additionally assume a strict tree structure with no loops or duplicate nodes, as our initial formulation does not include loopy belief propagation.

Each node $i$ in the tree maintains a probability distribution $P_i(x)$ over its true expected reward value. Inference of interior node probability distributions begins at the leaves and propagates up to the root node. Before any trials have been performed, leaf nodes are initialized to conjugate prior distributions that are appropriate to the leaf node reward distributions. For example, for 0/1 payoffs at the leaf nodes, we may use beta function priors $x^{\alpha-1}(1-x)^{\beta-1}/B(\alpha,\beta)$ where $\alpha$ and $\beta$ are effectively the number of prior wins and losses. When trials at leaf nodes are performed, the results are combined with priors in the standard way to compute posterior distributions. Again in the 0/1 case, after $W$ sample wins and $L$ sample losses, we can obtain the posterior by adding $W$ to $\alpha$ and $L$ to $\beta$ in the above expression. To propagate posteriors up the tree, we represent them either numerically, or by a Gaussian with matching mean and standard deviation, as detailed in section 4.1.

The distributions of leaf nodes propagate upward to parent, grandparent, etc., nodes. Each interior node computes an *extremum distribution* over its child node distributions, depending on its MAX or MIN type. The extremum distribution of a set of independent random variables $\{x_1, x_2, \ldots, x_K\}$ is a well-known concept in statistics. The MAX distribution is given by

$$P_{max}(X) = \sum_i P_i(X) \prod_{j \neq i} C_j(X) \quad (2)$$

where $C_j$ is the CDF of $P_j$. A more convenient calculation is to first compute the parent CDF $C_{max}(X) = \int P_{max}(X)dX$ as the product of child CDFs: $C_{max}(X) = \prod_i C_i(X)$ and then differentiate $C_{max}$ to obtain $P_{max}$. (For MIN nodes we compute $\prod_i (1 - C_i(X)) = 1 - C_{min}(X)$.) The parent MAX/MIN distributions are computed either by simple numeric integration, or by analytic Gaussian approximation as described below.

Having computed distributions for the interior nodes, we now consider distributional analogs of existing distribution-free sampling formulae. For our initial studies we have focused specifically on Bayesian analogs of UCT, due to its simplicity and practical success in applications such as Go. (However, Bayesian analogs of BAST etc. can also be easily devised.) We propose two modified versions of UCB1 (the sampling formula in UCT) to descend the tree and choose where to sample next. The first version (Bayes-UCT1) simply replaces the average reward $\bar{r}_i$ of child node $i$ by $\mu_i$, the mean of $P_i$:

$$\text{Bayes-UCT1: maximize } B_i = \mu_i + \sqrt{2 \ln N / n_i} \quad (3)$$

We strongly believe that equation 3 constitutes a strict improvement over UCT if the independence assumption and leaf node priors are correct, since the posterior mean $\mu_i$ provides a more accurate estimator of true node value than the average reward $\bar{r}_i$.

Our second sampling formula (Bayes-UCT2) additionally replaces the $1/\sqrt{n_i}$ factor in the exploration term by $\sigma_i$, the square root of the variance of $P_i$:

$$\text{Bayes-UCT2: maximize } B_i = \mu_i + \sqrt{2 \ln N} \sigma_i \quad (4)$$

Equation 4 is motivated by the central limit theorem result $\sigma_i^2 \sim 1/n_i$ in the simple bandit case, and by the compelling intuitive notion from Interval Estimation that sampling according to expected value plus expected uncertainty provides effective tradeoff of exploration vs. exploitation. Other uncertainty measures such as the credible interval may be even more effective, but using $\sigma$ provides the simplest approach to begin investigations of uncertainty-based sampling using distributional information. We provide sketches of rigorous convergence proof for equation 4 as well as more general off-policy convergence below.

### 3.1 CONVERGENCE PROOFS

We provide only proof sketches here; full proofs will be presented elsewhere.

*Lemma 1:* Consider a bandit arm leaf node which generates 0/1 payoffs at a true win rate of $\mu^*$ such that $0 < \mu^* < 1$. Assume a prior probability distribution $P_{prior}(p)$ which is strictly positive on $(0, 1)$ and which has bounded derivatives. Then as the number of samples $n \to \infty$, the posterior distribution $P_{post}(p)$ converges to $\delta(p - \mu^*)$ with probability 1.

*Proof sketch:* Write the posterior by Bayes' theorem, and differentiate $\log(P_{post})$ to show that it has a peak at $\mu^*$ plus a correction term $\sim O(1/\sqrt{n})$. The second derivative is $-n/(\mu^*(1 - \mu*)) + O(\sqrt{n})$ so that the posterior approaches a Gaussian with mean $\mu^*$ and variance $\mu^*(1 - \mu^*)/n$ as $n \to \infty$.

*Theorem 1:* (Off-policy convergence) Consider a fixed finite-sized bandit tree with binary reward leaf nodes and priors as per Lemma 1. Assume no two sibling nodes have exactly identical minimax payoff rates. Then for any sampling policy $\Pi$ (e.g., uniform random sampling) that samples all leaf nodes an unbounded number of times as the total number of samples $N \to \infty$, all interior node posterior distributions converge with probability 1 to delta functions at the minimax optimal values.

*Proof Sketch:* Show that if a parent's child nodes all converge to delta functions at correct minimax values, then the parent's distribution converges to the distribution of the best child according to MIN or MAX, due to finite separation of child values. By induction from leaf node convergence of Lemma 1, all interior nodes converge. (This also applies under our Gaussian MIN/MAX approximation framework of section 4.1 since the approximation error vanishes as the child variances $\to 0$.)

*Theorem 2:* (On-policy convergence) For the sampling policy specified in equation 4, all interior nodes converge with probability 1 to correct minimax optimal point values.

*Proof Sketch:* First show that finitely sampled leaf nodes have finite non-vanishing variance, and then show that a parent's variance is non-vanishing if its child node distributions have non-vanishing variance. By induction, all finitely sampled interior nodes have finite variance. Now suppose an interior node remains unsampled while its parent and all of its siblings receive infinite samples. In this case, the node's exploration term in equation 4 will grow without bound, while the bounds of the siblings all converge to their true means; hence the node must eventually receive more samples. By induction, all leaf nodes receive unbounded samples, so that Theorem 1 applies. (This also applies under Gaussian approximation since the approximated variance is strictly positive given non-vanishing child distribution variances.)

# 4 IMPLEMENTATION DETAILS

## 4.1 GAUSSIAN APPROXIMATION

Our primary approach to representing probability distributions is to approximate them with Gaussians, using recent advances in fast analytic computation of max/min of Gaussians to propagate inference in the tree. There are two potential major advantages of this approach: it may run orders of magnitude faster than numeric methods, plus it can also take correlations between Gaussians into account. Specifically, if a correlation model $\rho_{ij}$ is available between $K$ sibling nodes, we can estimate the correlation between the max/min distribution of any pair of siblings, and any third sibling. While we do not make use of this property in our test domain, it is likely to play an important role in further development of Bayesian MCTS methods.

The potential disadvantage of Gaussian approximation is that approximation errors could arise due to two sources. First, the leaf node distributions may have very few trials and may have certain priors (such as a uniform prior) that are poorly approximated by Gaussians. Secondly, it is known that the Gaussian family is not closed under the max operation, and making such an approximation may introduce further error. We address this issue below. In the uncorrelated case, the largest approximation errors occur when the input Gaussians have very similar means but very different widths. Fortunately, bandit sampling algorithms act to prevent this from happening (at least for the dominant Gaussians contributing most to the max distribution) as we explain later in section 5.2.

Closed-form analytic expressions for the mean $\mu$ and variance $\sigma^2$ of the maximum distribution of two Gaussian random variables were first calculated by Clarke in [7]. Let $\mu_1, \mu_2$ denote means and $\sigma_1, \sigma_2$ denote the standard deviations of two input Gaussians with correlation coefficient $\rho$. Expanding the moment generating function yields the following expressions:

$$\begin{aligned} \mu &= \mu_1 \Phi(\alpha) + \mu_2 \Phi(-\alpha) + \phi(\alpha)\sigma_m \\ \sigma^2 &= \left(\mu_1^2 + \sigma_1^2\right)\Phi(\alpha) + \left(\mu_2^2 + \sigma_2^2\right)\Phi(-\alpha) + \\ &\quad (\mu_1 + \mu_2)\sigma_m\phi(\alpha) - \mu^2 \end{aligned}$$

where $\sigma_m = \sigma_1^2 - 2\rho\sigma_1\sigma_2 + \sigma_2^2$, $\alpha = (\mu_1 - \mu_2)/\sigma_m$, $\phi()$ denotes a standard Gaussian PDF with zero mean and unit variance, and $\Phi()$ denotes the CDF of $\phi()$.

We reduce the computation time by restructuring the equations above to yield:

$$\mu = \mu_2 + \sigma_m F_1(\alpha); \ \sigma^2 = \sigma_2^2 + (\sigma_1^2 - \sigma_2^2)\Phi(\alpha) + \sigma_m^2 F_2(\alpha)$$

where $F_1(\alpha) = \alpha\Phi(\alpha) + \phi(\alpha)$, and $F_2(\alpha) = \alpha^2\Phi(\alpha)(1 - \Phi(\alpha)) + (1 - 2\Phi(\alpha))\alpha\phi(\alpha) - \phi(\alpha)^2$. We implement 1-d

lookup tables for $F_1()$, $F_2()$ and $\Phi()$, thereby reducing the computation of $\mu$ and $\sigma^2$ to about a dozen floating point operations.

Extending Clarke's work, Sinha et al. [17] recently studied the error of approximating max/min of a set of $K$ input Gaussians by iterative pairwise combination: two of the input Gaussians are combined and approximated by a Gaussian, which is then combined with a third Gaussian, and so on, until all input Gaussians have been combined. The error of the overall result may depend significantly on the order of combination. Sinha et al. propose a stagewise minimum-error combining scheme using a 3-d error lookup table which estimates the error $E_{ij}$ of combining any input pair $(i, j)$ based on three independent quantities $\rho_{ij}$, $\sigma_i/\sigma_j$ and $\alpha_{ij}$ as defined above. At each stage, all pairwise errors $E_{ij}$ are evaluated and the pair with minimum error is combined.

We have implemented the above min-error combining scheme by computing the errors of combining each of the $K$ choose 2 pairs and storing them in a heap. We then combine the pair with minimum error, update the combining errors, and iterate until done. This algorithm was found in [17] to give favorable approximation errors, but takes $O(K^2 logK)$ operations and $O(K^2)$ space to store the heap. We have also implemented a much simpler $O(K)$ random-order combining scheme which appears to perform just as well as min-error combining in our bandit-tree domain, while running much faster. It appears that bandit-style sampling may generally avoid the worst-case Gaussian approximation error scenarios, and we discuss evidence for this in section 5.

We compare our Gaussian framework with an alternative numeric representation scheme in which distributions are represented exactly (to within grid point resolution) and inference propagates by numeric integration. Specifically, we use $G = 1000$ grid points to represent PDFs, and we compute CDFs by numeric integration using the trapezoid rule, with error $\sim O(1/G^2)$. This provides a reference point of essentially exact inference for our Gaussian approximation method, but is too slow to be practically feasible.

### 4.2 EMPIRICAL TESTBED

The testbed for our initial studies of Bayes-UCT is an idealized bandit-tree simulator that constructs a minimax tree of depth $D$ starting from a root node, in which the leaf nodes are ordinary bandit arms that give binary 0/1 payoffs at fixed probabilities $\{p_i\}$ of arm $i$ returning 1. We assign independent random values $\{p_i\}$ over the unit interval, so that the arms are uncorrelated. We choose either uniform random values, or Gaussian distributed values with mean 0.5 and standard deviation 0.1. This provides an interesting test as we run Bayes-UCT using a uniform prior: in the former case, the prior is correct, while in the latter case, the assumed prior differs substantially from the correct prior. (Note that the uniform prior can be represented by a beta distribution with $\alpha = \beta = 1$ and can be easily combined with sample results to estimate posterior beta distributions, as explained previously.)

Each interior node in the tree typically has a common width $W$, although we have also run experiments where a node's width is set by uniform random choice between a minimum and maximum width. We assume that the tree topology in the UCT and Bayes-UCT searches is static, i.e., all tree nodes are already contained in the UCT representation, having been added either by sampling or by prior modeling, and no node additions occur during the searches.

## 5 RESULTS

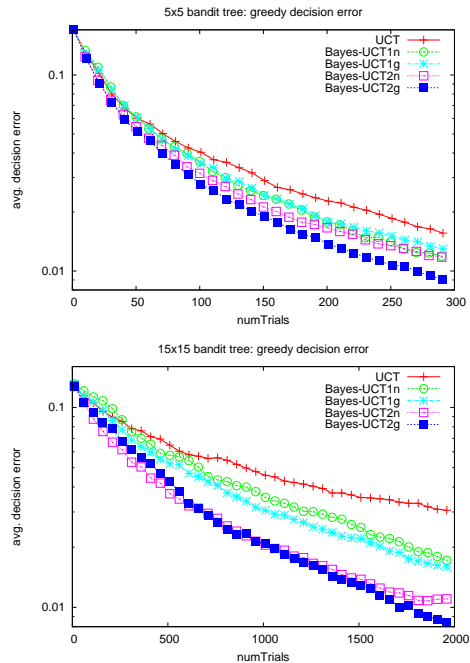

Figure 1: Average greedy decision error in 2-ply bandit tree. (a) width $W = 5$. (b) $W = 15$.

Figure 1(a) compares UCT with our Bayes-UCT1 and Bayes-UCT2 algorithms in a simple 2-ply minimax bandit tree topology containing a root MAX node with five MIN children, each of which has five children that are simple bandit arms. We assigned uniform random leaf node payoff rates so that the Bayes-UCT uniform prior is correct. As a function of number of simulator trials, we plot the average "greedy" decision error, i.e., the true loss of the top-level move with highest estimated mean value. The plot averages results over 3000 such bandit trees, with a single run performed for each tree. We duplicate the 3000 random trees for each algorithm, so that pairwise comparisons of algorithms have much higher significance than using independent random trees for each algorithm. We see in figure 1(a) that our Bayes-UCT1n and Bayes-UCT1g algorithms ("n" denotes numeric and "g" denotes Gaussian representation) initially perform slightly worse than UCT, but achieve clearly lower errors after approximately 50 trials. The Bayes-UCT2 algorithms outperform their Bayes-UCT1 counterparts, both initially and asymptotically, with Bayes-UCT2g achieving clearly the lowest error among all algorithms.

Figure 1(a) confirms our intuition that Bayes-UCT1 should outperform UCT by virtue of having more accurate node value estimates. (However, we don't have a clear understanding of why Bayes-UCT1 appears to underperform UCT initially.) It also confirms the expected improvement in Bayes-UCT2 from using a more accurate estimate of node uncertainty in the exploration term. However, given the potential errrors in Gaussian approximation, we were quite surprised that the Gaussians methods outperform their numeric counterparts! (Possible explanations are given below.)

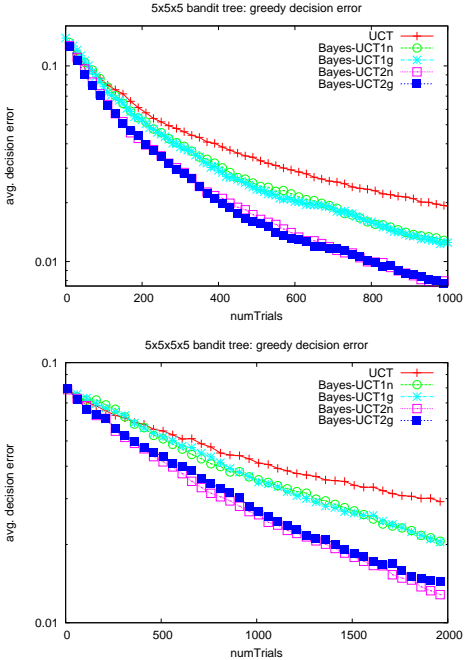

Figure 2: Average greedy decision error in width $W = 5$ bandit trees. (a) depth $D = 3$. (b) $D = 4$.

We also expected that UCT value estimates would be progressively less accurate relative to the Bayes-optimal estimates as the tree increases in either width or depth, and thus Bayes-UCT should obtain greater outperformance over UCT as either increases. This intuition turned out to be correct, as we see in figure 1(b), which shows results of similar experiments in 15x15 bandit trees. The gap between UCT and our methods has widened, and these plots confirm that: (a) Bayes-UCT1 slightly underperforms UCT very early in the simulation; (b) Bayes-UCT2 outperforms Bayes-UCT1; (c) Gaussian outperforms numeric representation.

Likewise, figures 2(a),(b) show widening outperformance of UCT as tree depth increases from $D = 2$ to $D = 3$ and $D = 4$. We still see that Bayes-UCT2 outperforms Bayes-UCT1. However, the Gaussian performance appears to weaken with increasing depth, so that at $D = 4$ Gaussian approximation now underperforms exact numeric representation.

We present further quantitative evidence comparing the number of simulated trials needed for UCT and Bayes-UCT2g to achieve an expected error level of 0.01. We have also included runs employing Gaussian distributed payoff rates, so that the uniform prior used in our Bayes-UCT implementation is incorrect. Results are shown in table 1. With a correct prior, there is strong evidence of scaling of Bayes-UCT2g advantage with both depth and width. Even with the incorrect prior, there is less compelling but still clear scaling. We also inadvertently discovered that with uniform payoffs, UCT is unable to converge in any feasible simulation time at $(D = 5, W = 5)$, whereas Bayes-UCT exhibits no signs of unusual convergence difficulty.

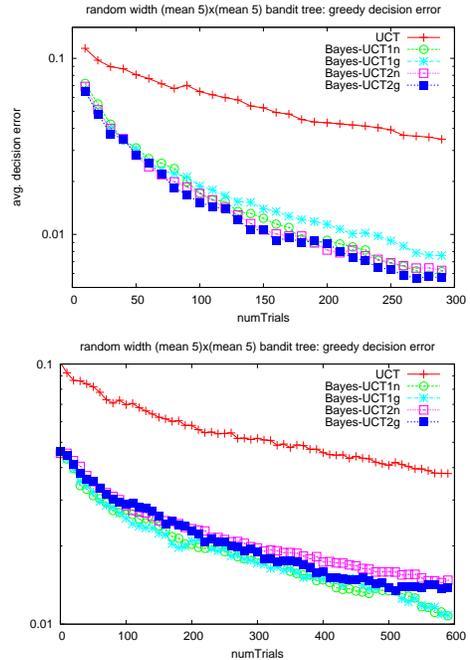

Figure 3: Average greedy decision error in 2-ply bandit tree, random width, mean=5. (a) Uniform payoff rates (correct prior). (b) Gaussian payoff rates (incorrect prior).

|  |  | uniform | payoffs |  | Gaussian | payoffs |  |
| --- | --- | --- | --- | --- | --- | --- | --- |
| Depth | Width | UCT | Bayes-UCT2g | Ratio | UCT | Bayes-UCT2g | Ratio |
| 5 | 5 | $> 10^6$ | 8330 | $> 100$ | 78890 | 33900 | 2.33 |
| 4 | 5 | 7330 | 2670 | 2.74 | 25170 | 11720 | 2.15 |
| 3 | 5 | 2050 | 780 | 2.63 | 7550 | 3890 | 1.94 |
| 2 | 5 | 480 | 290 | 1.66 | 1990 | 1240 | 1.60 |
| 2 | 10 | 2390 | 830 | 2.88 | 10650 | 4310 | 2.47 |
| 2 | 15 | 6440 | 1700 | 3.79 | 26240 | 8700 | 3.02 |
| 2 | 20 | 13390 | 3090 | 4.33 | 49210 | 15650 | 3.14 |

Table 1: Number of trials to achieve 0.01 average error rate in various bandit tree topologies. The ratio of UCT trials to Bayes-UCT2g trials increases with both width and depth of the bandit tree.

The final hypothesis which we have investigated is that UCT should have particular trouble in trees with non-uniform width, due to neglecting width either in the expected node values or in the exploration term. To study this, we modified the simulator to generate random-width trees, where each node is assigned a uniform random number of child nodes between a min and max number of allowable children. We then ran an experiment using two-ply random-width bandit trees with mean width around 5 at each level. (The bottom-level widths ranged between 1 and 10; top level width ranged between 2 and 10 so that the top level always had a nontrivial decision problem.) These experiments yielded our most striking results, shown in figure 3. Bayes-UCT massively outperforms UCT, particularly when it has a correct prior, but there is still very substantial improvement over UCT even using an incorrect prior.

### 5.1 SIMULATION SPEED

Table 2 gives measures of UCT and Bayes-UCT simulation speed in a few representative experiments. Since our stochastic "playouts" have negligible cost, Bayes-UCT obviously runs much slower (although within an order of magnitude) than UCT. To model expected behavior in an application like Go with heavyweight playouts, we have computed adjusted figures assuming a playout simulation time of 0.1 msec per trial[1]. In such scenarios the Bayes-UCT speed is within a factor of two of UCT and may be more than compensated by reduction in required number of trials.

### 5.2 BANDIT TREE INSIGHTS

Herein we provide insights we have gleaned in developing an understanding of how Bayes-UCT can perform so well compared to UCT. One key factor seems to be that Bayes-UCT calculates much more accurate node value estimates (given a correct prior, these are in fact Bayes-optimal estimates). We have verified in the 5x5 bandit tree experiment of figure 1(a) that Bayes-UCT's node estimates are much more accurate than the simple UCT estimates. This can be seen in figure 4(a), where we plot the average absolute error of all top-level node value estimates as a function of how many times they have been sampled. Every time a top-level node is sampled, we recompute the probability distribution and mean, note the absolute difference between mean and true value, and bin the data point according to total trials performed by the top-level node. The graph shows that we improve top-level node value accuracy over UCT by a factor of $\sim 3-5$. However, estimation accuracy is not the entire story, as we see from a further experiment shown in figure 4(b). In this experiment we devised a hybrid algorithm in which top-level preferences were computed based on Bayesian estimates, but actual sampling decisions were performed using UCT. The performance was substantially worse than full Bayes-UCT, showing that Bayes-UCT derives significant benefit from better sampling decisions in addition to having more accurate value estimates.

The remaining major puzzle for us was how the Gaussian approximation could do so well relative to exact numeric distributions. It turns out that the estimated first moments using Gaussians lie extremely close to the essentially exact estimates obtained from numeric integration. In all instances that we have observed, the difference between numeric mean and Gaussian mean is always $\sim 0.01$ or less. Moreover, there is no loss of accuracy in predicting the node's true value, as we see comparing Gaussian vs. numeric accuracy in figure 4(a). Our takeaways from these observations are: (a) Gaussian approximation is a lot better than we realized at estimating the mean of an extremum distribution (as opposed to its exact shape, which can have quite large errors); (b) due to the nature of bandit-based sampling, we tend not to encounter the large-error cases, which occur when the input Gaussians have nearly equal means and very different sigmas. If this ever were to happen, UCB-type sampling algorithms would devote many more samples to the wider input and shrink it to match the narrower input.

---

[1] In competitive $19 \times 19$ Go programs, playout simulation times of 0.5-1 msec are common, which would make our results look even more favorable. We chose a more conservative figure of 0.1 msec, corresponding to typical playout simulation time in $9 \times 9$ Go.

| Depth | Width | UCT-raw | UCT-adj | Bayes-UCT2g-raw | Bayes-UCT2-adj |
|---|---|---|---|---|---|
| 5 | 5 | 122.0 | 9.24 | 14.71 | 5.95 |
| 2 | 5 | 188.1 | 9.50 | 27.12 | 7.31 |
| 2 | 20 | 59.6 | 8.56 | 7.30 | 4.22 |

Table 2: Simulation speed (thousands of trials/sec); "raw" assumes zero-cost playouts and "adj" assumes 0.1 msec playouts.

We have similarly scrutinized the differences between numeric and analytic estimates of sigma values. It turns out the analytic estimates are very close to the numeric values, except in one particular case where the mean is close to 0 or 1, and the distribution is highly skewed. In this case the analytic sigmas are systematically much higher than the correct values. We argue that this "error" is actually of benefit to our Gaussian method! By setting sigma too high, the Gaussian is capturing more of the tail mass of the distribution due to higher-order moments. With so-called "fat tailed" distributions it is desirable to explore more than dictated by the true second moment of the distribution.

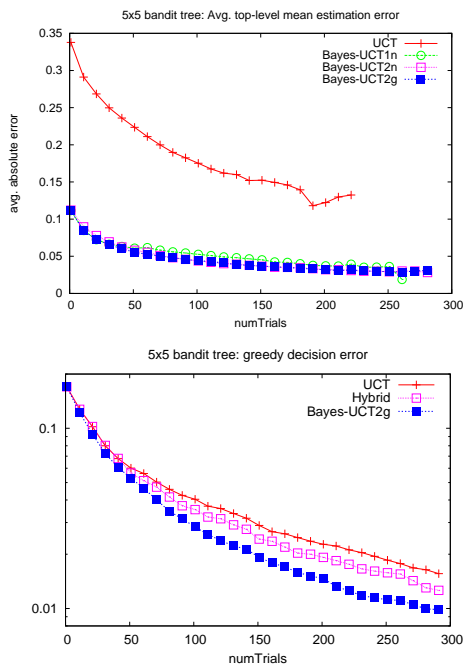

Figure 4: (a) Average error in estimated mean of top-level nodes. (b) Performance of Bayesian selection of top-level moves in a "Hybrid" algorithm where samples are guided by normal UCT.

## 6 RELATED WORK

Several extensions of UCT are known in the literature; most use the core UCT search strategy but provide alternative upper-bound functions to guide node selection. In many ways Bayes-UCT is most similar to Flat-UCB1 [8]. Flat-UCB1 uses the standard UCB1 bound function at terminal nodes, but the bounds for interior nodes are computed as the maximum value of all the bounds for the node's children. Bayes-UCT extends this approach by both using distributions and by including an exploration term for interior nodes.

BAST and HOO are extensions of UCT that provide better asymptotic regret for smooth trees [8, 5]. BAST and HOO use an alternative bound function that combines UCB1-PAC(b) with a smoothing term $\delta_d$ that represents the smoothness of the local tree. HOO uses the same principles as BAST but its bound function is adapted to solve continuous optimization problems. The applicability of BAST to Go and similar problems is questionable as Go does not exhibit the local smoothness BAST is designed to exploit. That is, $\delta_d$ does not vary with depth but instead has the constant value $\delta_d = 1$ for all nodes for which the game is not already decided.

UCB1-NORMAL [2] is similar to Bayes-UCT2 in that both formulas replace the $1/\sqrt{n}$ factor of UCB1 with the sample variance. The same paper proposes UCB1-TUNED which includes (without explicit justification) a separate exploration term on the variance. The authors demonstrate that UCB1-TUNED outperforms UCB1 for simple bandit problems, but the effectiveness of UCB1-NORMAL is not evaluated. Our results showing advantages of Bayes-UCT2 over Bayes-UCT1 suggest that similar advantages might be obtained by UCB1-NORMAL over standard UCT.

Our work is a natural evolution of probabilistic minimax search [15, 3] in applying it to MCTS search where a static evaluation function is not needed. This is important for computer Go where the design of effective evaluation functions has been elusive.

## 7 CONCLUSIONS

We presented an original Bayesian formulation of Monte-Carlo Tree Search, appropriate for independent

tree nodes, offering a number of theoretical advantages, including on-policy as well as more robust off-policy convergence guarantees. Empirically, we obtained some interesting and surprising insights by testing our methods in an idealized bandit-tree domain. We were encouraged to find significant and scalable improvements over UCT, even using an incorrect prior, and especially in the case of non-uniform tree widths. We were also surprised that our Gaussian approximation framework turned out to be more accurate and worked much better than we originally anticipated.

One immediate next step in our future work will be to compare BAST with its Bayesian counterparts, much as we have done here for UCT. BAST already incorporates simple min/max operations in its bound computations, and is designed to scale better with depth that UCT, so it will be interesting to see whether Bayes-BAST can obtain the same magnitude of improvements that we saw over UCT. We are also interested in further explorations of issues arising from non-uniform tree width. In this regard, it will be interesting to compare Bayes-UCT with PAC-UCB [1], an extension of UCB which does include the number of bandit arms in estimating upper confidence bounds. A third important direction for practical implementations is to develop incremental versions of the Gaussian combining schemes studied here; this could yield an order of magnitude speedup and eliminate most of the speed advantage that UCT has over Bayes-UCT.

However, the most important direction for future practical use of Bayesian MCTS is to develop methods for automatically modeling the correlation of sibling nodes in a given domain. The Gaussian min/max approximation framework does allow correlated input Gaussians, but we are unaware of any methods that can automatically estimate the appropriate correlation to use in a given MDP/game state. This appears to be the main challenge in applying Bayes-UCT to computer Go, as the use of crude correlation models (e.g., a single global correlation coefficient) has not shown an advantage over UCT. We are currently investigating several possible machine learning approaches to modeling correlations in Go and other challenging single-agent planning problems.

## Acknowledgements

The authors thank Mark Wegman, Jeff Kephart, David Silver, Csaba Szepesvari, and several anonymous reviewers for numerous helpful comments and suggestions. This work was supported in part under the DARPA GALE project, contract No. HR0011-08-C-0110.